\documentclass[preprint,12pt]{elsarticle}

\usepackage{graphicx}
\usepackage{amsmath}
\usepackage{amssymb}
\usepackage{algorithmic}
\usepackage{algorithm}
\usepackage{subfigure}
\usepackage{array}

\newcommand{\bfd}{{\textbf{d}}}
\newcommand{\bfs}{{\textbf{s}}}
\newcommand{\bfx}{{\textbf{x}}}

\newcommand{\bfw}{{\textbf{w}}}

\journal{Neural Networks}

\begin{document}

\begin{frontmatter}

\title{A novel multivariate performance optimization method based on sparse coding and hyper-predictor learning}

\author[X]{Jiachen Yang}

\author[X]{Zhiyong Ding}

\author[X]{Fei Guo\corref{cor1}}
\ead{gfjy001@yahoo.com}

\author[X]{Huogen Wang}

\author[Y]{Nick Hughes}
\ead{Nick.Hughes1@yahoo.com	}

\address[X]{School of Electronic Information Engineering, Tianjin University, Tianjin 300072, China}

\address[Y]{College of Information Science and Technology, University of Nebraska Omaha, Omaha, NE 68182,United States}

\cortext[cor1]{Corresponding author: Fei Guo. E-mail: gfjy001@yahoo.com. Tel: +86-13821820218.}

\begin{abstract}
In this paper, we investigate the problem of optimization multivariate performance measures, and propose a novel algorithm for it. Different from traditional machine learning methods which optimize simple loss functions to learn prediction function, the problem studied in this paper is how to learn effective hyper-predictor for a tuple of data points, so that a complex loss function corresponding to a multivariate performance measure can be minimized. We propose to present the tuple of data points to a tuple of sparse codes via a dictionary, and then apply a linear function to compare a sparse code against a give candidate class label. To learn the dictionary, sparse codes, and parameter of the linear function, we propose a joint optimization problem. In this problem, the both the reconstruction error and sparsity of sparse code, and the upper bound of the complex loss function are minimized. Moreover, the upper bound of the loss function is approximated by the sparse codes and the linear function parameter. To optimize this problem, we develop an iterative algorithm based on descent gradient methods to learn the sparse codes and hyper-predictor parameter alternately. Experiment results on some benchmark data sets show the advantage of the proposed methods over other state-of-the-art algorithms.
\end{abstract}


\end{frontmatter}

\section{Introduction}

In traditional machine learning methods, we usually use a loss function to compare the true class label of a data point against its predicted class label. By optimizing the loss functions over all the training set, we seek a optimal prediction function, named a classifier \cite{Micheloni201281,Roy2013113,Kang201439,Bhuyan2011,Wang2012Multiple}. For example, in support vector machine (SVM), a hinge loss function is minimized, and in linear regression (LR), a logistic loss function is used \cite{Pragidis201522,Couellan20154284,Siray2015217,Patil2015349}. However, when we evaluate the performance of a class label predictor, we usually consider a tuple of data points, and use a complex multivariate performance measure over the considered tuple of data points, which is different from the loss functions used in the training procedure significantly \cite{Joachims2005377,Walker201155,Zhang2011814,Zhang20123623,Mao20132051}. For example, we may use area under receiver operating characteristic curve (AUC) as a multivariate performance measure to evaluate the classification performance of SVM. Because SVM class label predictor is trained by minimizing the loss functions over training data points, it cannot be guaranteed to minimize the loss function corresponding to AUC. Many other multivariate performance measures are also defined to compare a true class label tuple of a data point tuple against its predicted class label tuple, and they can also be used for different machine learning applications. Some examples of the multivariate performance measures are as F-score \cite{Zemmoudj2014371,Gao2014}, precision-recall curve eleven point (PRBEP) \cite{Boyd2013451,Lopes2014322}, and Matthews correlation coefficient  (MCC) \cite{Kumari2015175,Shepperd2015106}. To seek the optimal multivariate performance measures on a given tuple of data points, recently, the problem of multivariate performance measure optimization is proposed. This problem is defined as a problem of learning a hyper-predictor for a tuple of data points to predict a tuple of class labels. The hyper-predictor is learned so that a multivariate performance measure used to compare the true class label tuple and the predicted class label tuple can be optimized directly.

\subsection{Related works}

Some methods have been proposed to solve the problem of multivariate performance measures. For example,

\begin{itemize}
\item Joachims \cite{Joachims2005377} proposed a SVM method to optimize multivariate nonlinear performance measures, including F-score, AUC etc. This method takes a multivariate predictor, and gives an algorithm to train the a multivariate SVM  in polynomial time for large classes so that the potentially non-linear performance measures can be optimized. Moreover, the translational SVM with hinge loss function can be treated as a special case of this method.

\item Zhang et al. \cite{Zhang2011814} proposed a smoothing strategy for multivariate performance score optimization., in particular PREBP and AUC. The proposed method combines Nesterov's accelerated gradient algorithm and the  smoothing strategy, and obtains an optimization algorithm. This algorithm converges to a given accurate solution in a limited number of iterations corresponding to the accurate.

\item Mao and Tsang \cite{Mao20132051} proposed a generalized sparse regularizer for multivariate performance measure optimization. Based on the this regularizer, a unified feature selection and general loss function optimization is developed. The formulation of the problem is solved by a two-layer cutting plane algorithm,  and the convergence is presented. Moreover, it can also be used to optimize the multivariate measures of multiple-instance learning problems.

\item Li et al. \cite{Li20131370} proposed to learn a nonlinear classifier for optimization of nonlinear and nonsmooth performance measures by  novel two-step approach. Firstly,  a nonlinear auxiliary classifiers with existing learning methods is trained,  and then it is adapted for specific performance measures. The classifier adaptation can be reduced to a quadratic program problem, similar to the method introduced in \cite{Joachims2005377}.

\end{itemize}

\subsection{Contributions}

In this paper, we try to investigate the usage of sparse coding in the problem of multivariate performance optimization. Our work is inspired by the work of multivariate performance optimization using multiple kernel learning proposed by Wang, et al. \cite{wang2015multiple}. The work in \cite{wang2015multiple} is a original contribution of major significance, because for the first time, it proposed to map the data into another space to learn a more effective predictor in the new space for multivariate performance measure optimization. Specifically, it uses multiple kernel learning \cite{wang2014effective} to map the input data to a new space, and then learns a new predictor to optimize the desired multivariate performance measure. Our work also follows this strategy, but our work uses sparse coding to map the original input data to a new sparse code space, instead of using multiple kernel learning. Moreover, our method also learns a new predictor in the new space to optimize the multivariate performance measure. Sparse coding is an important and popular data representation method, and it represent a given data point by reconstructing it with regard to a dictionary \cite{Li20151254,AlShedivat20141665,Wang20141630,Wang20133249,wang2015representing}. The reconstruction coefficients are imposed to be sparse, and used as a new representation of the data point. Sparse coding has been used widely in both machine learning and computer vision communities for pattern classification  problems. For example, Mairal et al. \cite{Mairal20091033} proposed to learn the sparse codes and a classifier jointly on a training set. However, the loss function used in this method is a traditional logistic loss. In this paper, we ask the following question: How can we learn the sparse codes and its corresponding class prediction function to optimize a multivariate performance measure? To answer this question, we propose a novel multivariate performance optimization method. In this method, we try to learn sparse codes from the tuple of training data points, and apply a linear function to match the sparse code tuple against a candidate class label. Based on the linear function, we design a hyper-predictor to predict the optimal class label tuple. Moreover, to the loss function of the desired multivariate performance measure is used to compare the prediction of the hyper-predictor and the true class label tuple, and minimized to optimize the multivariate performance measure. The contributions of this paper are of two folds:

\begin{enumerate}
\item We proposed a joint model of sparse coding and multivariate performance measure optimization. We learn both the sparse codes and the hyper-predictor to optimize the desired multivariate performance measure. The input of the hyper-prediction function is the tuple of the sparse codes, and the output is a class label tuple, which is further compared the to the true class label tuple by a multivariate performance measure.  A joint optimization problem is constructed for this problem. In the objective function of the optimization problem, both the reconstruction error and the sparsity of the sparse code are considered. Simultaneously, the multivariate loss function of the multivariate performance function is also included in the objective. The multivariate loss function may be very complex, and even does not have a close form, thus it is difficult to optimize it directly. We seek its upper bound, and approximate is as a linear function of the hyper-predictor function.

\item We proposed a novel iterative algorithm to optimize the proposed problem. We adapt the alternate optimization strategy, and optimize the sparse code, dictionary and the hyper-predictor function alternately in an iterative algorithm. Both sparse codes and hyper-predictor parameters are learned by gradient descent methods, and the dictionary is learned by Lagrange multiplier method.
\end{enumerate}

\subsection{Paper organization}

This paper is organized as follows. In section \ref{sec:method}, we introduce the proposed multivariate performance measure optimization method. In section \ref{sec:exp}, the proposed method is evaluated experimentally and compared to state-of-the-art multivariate performance measure optimization methods. In section \ref{sec:conclud}, the paper is concluded with future works.

\section{Proposed method}
\label{sec:method}

In this section, we introduce the proposed method. We first model the problem with an optimization problem, then solve it with an iterative optimization strategy, and finally develop an iterative algorithm based on the optimization results.

\subsection{Problem formulation}

\begin{figure}
  \centering
  \includegraphics[width=0.9\textwidth]{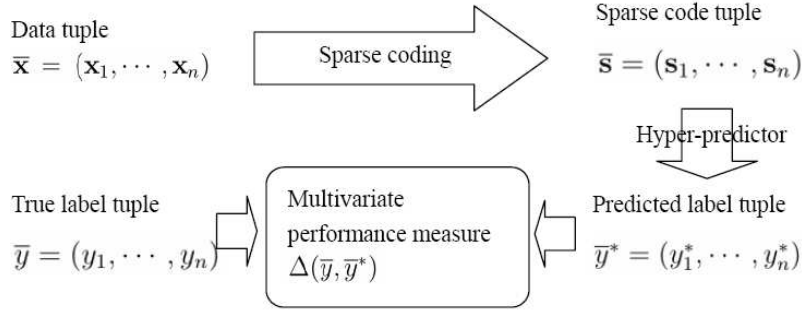}\\
  \caption{Flowchart of the proposed learning framework.}
  \label{fig:FigChatI150604}
\end{figure}

Suppose we have a tuple of $n$ training data points, $\overline{\bfx} = (\bfx_1, \cdots, \bfx_n)$, and its corresponding class label tuple is denoted as $\overline{y} = (y_1, \cdots, y_n)$, where $\bfx_i\in \mathbb{R}^d$ is the $d$-dimensional feature vector of the $i$-th training data point, and $y_i \in \{+1, -1\}$ is the binary label of the $i$-th training data point. We can use a machine learning method to predict the class label tuple, $\overline{y}^* = (y_1^*, \cdots, y_n^*)$, where $y_i^*$ is the predicted class label of the $i$-th data point. A multivariate performance measure, $\Delta(\overline{y},\overline{y}^*)$, is defined to compare a predicted class label tuple $\overline{y}^*$ of a data point tuple against its true class label tuple $\overline{y}$. To learn a hyper-predictor  
to map a data point tuple $\overline{\bfx}$ to a optimal class label tuple $\overline{y}^*$, we should learn it to minimize a desired pre-defined multivariate performance measure, $\Delta(\overline{y},\overline{y}^*)$. The proposed learning framework is shown in the flowchart in Fig. \ref{fig:FigChatI150604}.

We propose to present the data points to their sparse codes by sparse coding method, and then use a linear hyper-predictor to predict the class label tuple. We consider the follow problems in the learning procedure,

\begin{itemize}
\item \textbf{Sparse coding of data tuple}: To represent the data points in the data tuple, we propose to reconstruct each data point in the data tuple by using a dictionary,

\begin{equation}
\label{equ:sparsecoding}
\begin{aligned}
\bfx_i \approx \sum_{j=1}^m s_{ij} \bfd_j = D \bfs_i, ~i=1,\cdots,n,
\end{aligned}
\end{equation}
where $\bfd_j \in \mathbb{R}^d$ is the $j$-th dictionary element of the dictionary, and $D = [\bfd_1, \cdots, \bfd_m]$ is the dictionary matrix with its $j$-th column as the $j$-th dictionary element, and $m$ is the number of the dictionary elements. $s_{ij}$ is the coefficient of the $j$-th dictionary element for the reconstruction of the $i$-th data point, and $\bfs_i = [s_{i1}, \cdots, s_{im}]^\top \in \mathbb{R}^m$ is the coefficient vector for the reconstruction of the $i$-th data point. We assume that for each data point, only a few dictionary elements are used, thus its coefficient should be sparse, and we also call it sparse code of the data point. To learn the dictionary and the sparse codes of the data tuple, we propose to minimize the reconstruction error and encourage the sparsity of the sparse codes, and the following optimization problem is obtained over the data tuple,

\begin{equation}
\label{equ:sparsecoding_object}
\begin{aligned}
\min_{D, \bfs_i|_{i=1}^n} ~
&\sum_{i=1}^n \left ( \left \| \bfx_i - D \bfs_i \right \|_2^2 + C_1 \|\bfs_i\|_1 \right ),\\
s.t.~
& \|\bfd_j\|_2^2 \leq c, \forall~j=1,\cdots, m.
\end{aligned}
\end{equation}
In the objective function, the first part of each term is the reconstruction error measured by squared $\ell_2$ norm, and the second part is the sparsity measured by the $\ell_1$ norm of $\bfs_i$. {$C_1$ is a tradeoff parameter to control the sparsity of $\bfs_i$. If we have a larger value of $C_1$, the learned $\bfs_i$ will be more sparse. The optimal value of this parameter can be selected by linear search or cross validation.}

\item \textbf{Learning of hyper-predictor}: We apply a linear function, $f(\overline{\bfs},\overline{y}')$,  to compare the tuple of sparse codes of the data tuple, $\overline{\bfs} = (\bfs_1, \cdots, \bfs_n)$, against a candidate class tuple, $\overline{y}' = (y_1', \cdots, y_n')$,

\begin{equation}
\label{equ:linear}
\begin{aligned}
f(\overline{\bfs},\overline{y}') =  \sum_{i=1}^n y_i' \bfw^\top \bfs_i,
\end{aligned}
\end{equation}
where $\bfw \in \mathbb{R}^m$ is the parameter vector of the function. Then we the candidate class label tuple $\overline{y}'$ which archives the largest response of $f(\overline{\bfs},\overline{y}')$ will be output as the optimal class label tuple,

\begin{equation}
\label{equ:hyper_predictor}
\begin{aligned}
\overline{y}^* =  \underset{\overline{y}'\in \mathcal{Y}}{\arg\max} f(\overline{\bfs},\overline{y}')
\end{aligned}
\end{equation}
where $\mathcal{Y} = \{+1,-1\}^n$ is the hyper-space of the candidate class label tuple. To learn the linear function parameter vector $\bfw$ for the hyper-predictor and the sparse codes, we propose to learn it by minimizing a loss function of a pre-defined multivariate performance measure, $\Delta(\overline{y}^*,\overline{y})$. To reduce the complexity of the linear function, we also propose to minimize the squared $\ell_2$ norm of the linear function parameter $\bfw$. Thus we propose the following optimization problem to learn $\bfw$,

\begin{equation}
\label{equ:obj_w}
\begin{aligned}
\min_{\bfw, \bfs_i|_{i=1}^n}
\left \{ \frac{C_2}{2}\|\bfw\|_2^2 +C_3 \Delta(\overline{y}^*,\overline{y}) \right \},
\end{aligned}
\end{equation}
{where $C_2$ and $C_3$ are other tradeoff parameters. $C_2$ is the weight of the model complexity penalty term, and a larger $C_2$ can leads to a simpler model. $C_3$  is the weight of the loss functions over the training data points, and a larger value of $C_3$ can lead the model to fit the training set better. The values of $C_2$ and $C_3$ can be selected by linear search of cross validation.} Direct minimization of $\Delta(\overline{y}^*,\overline{y})$ is difficult, thus we seek its upper bound and minimize its upper bound to optimize $\Delta(\overline{y}^*,\overline{y})$.

\begin{description}
\item[Theorem 1] The upper bound of $\Delta(\overline{y}^*,\overline{y})$ can be obtained as follows,

\begin{equation}
\label{equ:upper_bound}
\begin{aligned}
\frac{1}{\sum_{\overline{y}'': \overline{y}''\in \mathcal{Y}}
\tau_{\overline{y}''}}
\sum_{\overline{y}'': \overline{y}''\in \mathcal{Y}}
\tau_{\overline{y}''}
F(\overline{y}'')
\geq \Delta(\overline{y}^*,\overline{y}),
\end{aligned}
\end{equation}
where

\begin{equation}
\label{equ:F}
\begin{aligned}
F(\overline{y}'') =
\sum_{i=1}^n (y_i'' - y_i) \bfw^\top \bfs_i + \Delta(\overline{y}'',\overline{y}),
\end{aligned}
\end{equation}
and

\begin{equation}
\label{equ:tau}
\begin{aligned}
\tau_{\overline{y}''} =
\left\{\begin{matrix}
1, & if~F(\overline{y}'')\geq F(\overline{y}'''), \forall~\overline{y}'''\in \mathcal{Y}\\
0, & otherwise.
\end{matrix}\right.
\end{aligned}
\end{equation}

\item[Proof] According to (\ref{equ:hyper_predictor}), since $\overline{y}^*$ achieves a maximum $f(\overline{\bfs},\overline{y}') $,  we have

\begin{equation}
\label{equ:upper1}
\begin{aligned}
&f(\overline{\bfs},\overline{y}^*) \geq f(\overline{\bfs},\overline{y}) \\
&\Rightarrow
f(\overline{\bfs},\overline{y}^*) - f(\overline{\bfs},\overline{y}) \geq 0\\
&\Rightarrow
f(\overline{\bfs},\overline{y}^*) - f(\overline{\bfs},\overline{y}) + \Delta(\overline{y}^*,\overline{y}) \geq
\Delta(\overline{y}^*,\overline{y}).
\end{aligned}
\end{equation}
Substituting (\ref{equ:linear}) to the left hand of (\ref{equ:upper1}), and according to the definition of function $F(\overline{y}'')$ in (\ref{equ:F}), we have

\begin{equation}
\label{equ:upper2}
\begin{aligned}
& f(\overline{\bfs},\overline{y}^*) - f(\overline{\bfs},\overline{y}) + \Delta(\overline{y}^*,\overline{y})  \\
&= \sum_{i=1}^n y_i^* \bfw^\top \bfs_i - \sum_{i=1}^n y_i \bfw^\top \bfs_i + \Delta(\overline{y}^*,\overline{y})\\
& =
\sum_{i=1}^n (y_i^* - y_i) \bfw^\top \bfs_i + \Delta(\overline{y}^*,\overline{y})\\
& = F(\overline{y}^*).
\end{aligned}
\end{equation}
Thus (\ref{equ:upper1}) can be rewritten as

\begin{equation}
\label{equ:upper2}
\begin{aligned}
F(\overline{y}^*) \geq
\Delta(\overline{y}^*,\overline{y}).
\end{aligned}
\end{equation}
To find the upper bound of $F(\overline{y}^*)$, we scan all the candidate class label tuples $y''\in \mathcal{Y}$, and seek the one or more candidates which can achieve the maximum $F(\overline{y}'')$, and we can see the maximum $F(\overline{y}'')$ is a upper bound of $F(\overline{y}^*)$,

\begin{equation}
\label{equ:upper3}
\begin{aligned}
\max_{y''\in \mathcal{Y}} F(\overline{y}'') \geq F(\overline{y}^*)
\end{aligned}
\end{equation}
Moreover, we also define a indicator $\tau_{\overline{y}''}$ for each $\overline{y}''$ to indicate if $\overline{y}''$ achieves the maximum $F(\overline{y}'')$, as in (\ref{equ:tau}). In this way, we can rewrite the left hand of (\ref{equ:upper3}) as follows,

\begin{equation}
\label{equ:upper4}
\begin{aligned}
\max_{y''\in \mathcal{Y}} F(\overline{y}'') =
\frac{1}{\sum_{\overline{y}'': \overline{y}''\in \mathcal{Y}}
\tau_{\overline{y}''}}
\sum_{\overline{y}'': \overline{y}''\in \mathcal{Y}}
\tau_{\overline{y}''}
F(\overline{y}'').
\end{aligned}
\end{equation}
Thus we have

\begin{equation}
\label{equ:upper5}
\begin{aligned}
\frac{1}{\sum_{\overline{y}'': \overline{y}''\in \mathcal{Y}}
\tau_{\overline{y}''}}
\sum_{\overline{y}'': \overline{y}''\in \mathcal{Y}}
\tau_{\overline{y}''}
F(\overline{y}'')
=
\max_{y''\in \mathcal{Y}} F(\overline{y}'')
\geq
F(\overline{y}^*)
\geq \Delta(\overline{y}^*,\overline{y}).
\end{aligned}
\end{equation}

\end{description}

Instead of minimizing $\Delta(\overline{y}^*,\overline{y})$, we minimize its upper bound in (\ref{equ:upper_bound}), and (\ref{equ:obj_w}) is turned out to

\begin{equation}
\label{equ:obj_loss}
\begin{aligned}
\min_{\bfw, \bfs_i|_{i=1}^n}
&\left \{ \frac{C_2}{2}\|\bfw\|_2^2 +
\frac{C_3 }{\sum_{\overline{y}'': \overline{y}''\in \mathcal{Y}}
\tau_{\overline{y}''}}
\sum_{\overline{y}'': \overline{y}''\in \mathcal{Y}}
\tau_{\overline{y}''}
F(\overline{y}'')\right .\\
&=
\left .
\frac{C_2}{2}\|\bfw\|_2^2 +
\frac{C_3}{\sum_{\overline{y}'': \overline{y}''\in \mathcal{Y}}
\tau_{\overline{y}''}}
\sum_{\overline{y}'': \overline{y}''\in \mathcal{Y}}
\tau_{\overline{y}''}
\left ( \sum_{i=1}^n (y_i'' - y_i) \bfw^\top \bfs_i + \Delta(\overline{y}'',\overline{y}) \right )
 \right \}.
\end{aligned}
\end{equation}

\end{itemize}

The overall optimization problem is obtained by combining both problems in (\ref{equ:sparsecoding_object}) and {(\ref{equ:obj_loss})},

\begin{equation}
\label{equ:obj_overall}
\begin{aligned}
\min_{D, \bfs_i|_{i=1}^n, \bfw}
&
\left \{
\sum_{i=1}^n \left ( \left \| \bfx_i - D \bfs_i \right \|_2^2 + C_1 \|\bfs_i\|_1 \right )
\vphantom{
\frac{C_2}{\sum_{\overline{y}'': \overline{y}''\in \mathcal{Y}}
\tau_{\overline{y}''}}
\sum_{\overline{y}'': \overline{y}''\in \mathcal{Y}}
\tau_{\overline{y}''}
\left ( \sum_{i=1}^n (y_i'' - y_i) \bfw^\top \bfs_i + \Delta(\overline{y}'',\overline{y}) \right )
}
\right .
\\
& \left .
+
\frac{C_2}{2}\|\bfw\|_2^2 +
\frac{C_3}{\sum_{\overline{y}'': \overline{y}''\in \mathcal{Y}}
\tau_{\overline{y}''}}
\sum_{\overline{y}'': \overline{y}''\in \mathcal{Y}}
\tau_{\overline{y}''}
\left ( \sum_{i=1}^n (y_i'' - y_i) \bfw^\top \bfs_i + \Delta(\overline{y}'',\overline{y}) \right )
 \right \}.\\
s.t.~
&
\|\bfd_j\|_2^2 \leq c, \forall ~j=1, \cdots, m.
\end{aligned}
\end{equation}
In this problem, we learn the dictionary, sparse codes, and the hyper-predictor parameter jointly.

\subsection{Problem optimization}

\begin{figure}
  \centering
  \includegraphics[width=0.5\textwidth]{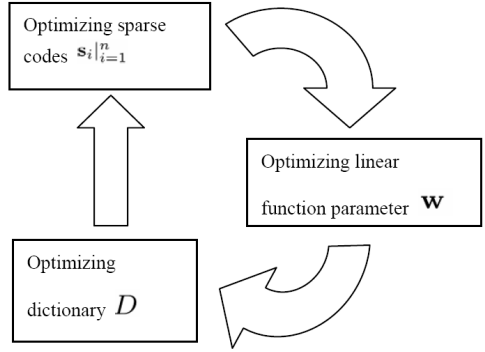}\\
  \caption{Flowchart of the alternate optimization strategy.}
  \label{fig:FigChartII150604}
\end{figure}

To optimize the problem in (\ref{equ:obj_overall}), we use the alternate optimization strategy. In an iterative algorithm, the variables are updated in turn. When the sparse codes are optimized, the linear function parameter and the dictionary are fixed. When the linear function parameter is optimized, the sparse codes and the dictionary are fixed. When the dictionary is optimized, the sparse codes and the linear function parameter is fixed. This strategy is shown in a flowchart in Fig. \ref{fig:FigChartII150604}.

\subsubsection{Optimization of sparse codes}

When we try to optimize the sparse codes, we fix the dictionary and the linear function parameter, and optimize the sparse codes one by one, i.e., when one sparse code $\bfs_i$ is considered, other sparse codes $\bfs_{i'}|_{i' \neq i}$ are fixed. Thus we turn the problem in (\ref{equ:obj_overall}) to the following optimization problem by only considering $\bfs_i$, and removing terms irrelevant to $\bfs_i$,

\begin{equation}
\label{equ:obj_s1}
\begin{aligned}
\min_{\bfs_i}
&
\left \{
\left \| \bfx_i - D \bfs_i \right \|_2^2 + C_1 \|\bfs_i\|_1
\vphantom{
\frac{C_2}{\sum_{\overline{y}'': \overline{y}''\in \mathcal{Y}}
\tau_{\overline{y}''}}
\sum_{\overline{y}'': \overline{y}''\in \mathcal{Y}}
\tau_{\overline{y}''}
\left ( \sum_{i=1}^n (y_i'' - y_i) \bfw^\top \bfs_i + \Delta(\overline{y}'',\overline{y}) \right )
}
\right .
\\
& \left .
+
\frac{C_3}{\sum_{\overline{y}'': \overline{y}''\in \mathcal{Y}}
\tau_{\overline{y}''}}
\sum_{\overline{y}'': \overline{y}''\in \mathcal{Y}}
\tau_{\overline{y}''}  (y_i'' - y_i) \bfw^\top \bfs_i
 \right \}.
\end{aligned}
\end{equation}
We rewrite the sparsity term in (\ref{equ:obj_s1}), $\|\bfs_i\|_1 $, as follows,

\begin{equation}
\label{equ:obj_s_norm}
\begin{aligned}
\|\bfs_i\|_1 = \sum_{j=1}^m |s_{ij}| = \sum_{j=1}^m \frac{s_{ij}^2}{|s_{ij}| }
= \bfs_i^\top diag \left (\frac{1}{|s_{i1}|}, \cdots, \frac{1}{|s_{im}|} \right ) \bfs_i,
\end{aligned}
\end{equation}
where $diag \left (\frac{1}{|s_{i1}|}, \cdots, \frac{1}{|s_{im}|} \right )$ is a diagonal matrix with its $j$-th diagonal element as $\frac{1}{|s_{ij}|}$. To make the objective function a smooth function, we fix the sparse code elements in the diagonal matrix as the elements of the previous iteration. Moreover, we note that $\tau_{\overline{y}''}$ is also a function of $\bfs_i$ as shown in (\ref{equ:tau}). We also first calculate by using sparse codes solved in the previous iteration, and then fix it when we consider $\bfs_i$ in the current iteration. In this way, (\ref{equ:obj_s1}) is changed to

\begin{equation}
\label{equ:obj_s2}
\begin{aligned}
\min_{\bfs_i}
&
\left \{
g(\bfs_i) =
\left \| \bfx_i - D \bfs_i \right \|_2^2 + C_1 \bfs_i^\top diag \left (\frac{1}{|s_{i1}^{pre}|}, \cdots, \frac{1}{|s_{im}^{pre}|} \right ) \bfs_i
\vphantom{
\frac{C_2}{\sum_{\overline{y}'': \overline{y}''\in \mathcal{Y}}
\tau_{\overline{y}''}^{pre}}
\sum_{\overline{y}'': \overline{y}''\in \mathcal{Y}}
\tau_{\overline{y}''}^{pre}
\left ( \sum_{i=1}^n (y_i'' - y_i) \bfw^\top \bfs_i + \Delta(\overline{y}'',\overline{y}) \right )
}
\right .
\\
& \left .
+
\frac{C_3}{\sum_{\overline{y}'': \overline{y}''\in \mathcal{Y}}
\tau_{\overline{y}''}^{pre}}
\sum_{\overline{y}'': \overline{y}''\in \mathcal{Y}}
\tau_{\overline{y}''}^{pre}  (y_i'' - y_i) \bfw^\top \bfs_i
 \right \},
\end{aligned}
\end{equation}
where $s_{ij}^{pre}$ is the $j$-th element of $\bfs_i$ solved in previous iteration, and $\tau_{\overline{y}''}^{pre}$ is $\tau_{\overline{y}''}$ calculated using previous solved $\bfs_i$ and $\bfw$. To seek its minimization, we update $\bfs_i$ by descending it to its gradient of the object $g(\bfs_i)$,

\begin{equation}
\label{equ:gradient_s}
\begin{aligned}
\bfs_i^{cur}  \leftarrow \bfs_i^{pre} - \eta \nabla g(\bfs_i) |_{\bfs_i = \bfs_i^{pre}},
\end{aligned}
\end{equation}
where $\bfs_i^{cur}$ is the sparse code updated in current iteration, $\bfs_i^{pre}$ is the sparse code solved in previous iteration, $\eta$ is the descent step, and $\nabla g(\bfs_i)$ is the gradient of $g(\bfs_i)$, which is defined as

\begin{equation}
\label{equ:obj_s3}
\begin{aligned}
\nabla g(\bfs_i)
&=
2 D^\top \left ( \bfx_i - D \bfs_i \right ) + 2 C_1  diag \left (\frac{1}{|s_{i1}^{pre}|}, \cdots, \frac{1}{|s_{im}^{pre}|} \right ) \bfs_i
\vphantom{
\frac{C_2}{\sum_{\overline{y}'': \overline{y}''\in \mathcal{Y}}
\tau_{\overline{y}''}^{pre}}
\sum_{\overline{y}'': \overline{y}''\in \mathcal{Y}}
\tau_{\overline{y}''}^{pre}
\left ( \sum_{i=1}^n (y_i'' - y_i) \bfw^\top \bfs_i + \Delta(\overline{y}'',\overline{y}) \right )
}
\\
&
+
\frac{C_3}{\sum_{\overline{y}'': \overline{y}''\in \mathcal{Y}}
\tau_{\overline{y}''}}
\sum_{\overline{y}'': \overline{y}''\in \mathcal{Y}}
\tau_{\overline{y}''}  (y_i'' - y_i) \bfw.
\end{aligned}
\end{equation}

\subsubsection{Optimization of linear function parameter}

By only considering $\bfw$ in  (\ref{equ:obj_overall}), fixing sparse codes, dictionary, and $\tau_{\overline{y}''}$ as results of previous iteration, and removing the terms irrelevant to $\bfw$, we turn (\ref{equ:obj_overall}) to

\begin{equation}
\label{equ:obj_w1}
\begin{aligned}
\min_{\bfw}
&
\left \{h(\bfw) =
\frac{C_2}{2}\|\bfw\|_2^2 +
\frac{C_3}{\sum_{\overline{y}'': \overline{y}''\in \mathcal{Y}}
\tau_{\overline{y}''}^{pre}}
\sum_{\overline{y}'': \overline{y}''\in \mathcal{Y}}
\tau_{\overline{y}''}^{pre}
\left ( \sum_{i=1}^n (y_i'' - y_i) \bfw^\top \bfs_i + \Delta(\overline{y}'',\overline{y}) \right )
 \right \}.
\end{aligned}
\end{equation}
To minimize this objective function, we update $\bfw$ by descending it to the gradient of $h(\bfw)$,

\begin{equation}
\label{equ:obj_w2}
\begin{aligned}
\bfw^{cur} \leftarrow \bfw^{pre} - \eta \nabla h(\bfw)|_{\bfw = \bfw^{pre}},
\end{aligned}
\end{equation}
where $\nabla h(\bfw)$ is the gradient of $h(\bfw)$, which is defined as

\begin{equation}
\label{equ:gradient_w1}
\begin{aligned}
\nabla h(\bfw) =
C_2\bfw  +
\frac{C_3}{\sum_{\overline{y}'': \overline{y}''\in \mathcal{Y}}
\tau_{\overline{y}''}^{pre}}
\sum_{\overline{y}'': \overline{y}''\in \mathcal{Y}}
\tau_{\overline{y}''}^{pre}
\sum_{i=1}^n (y_i'' - y_i)  \bfs_i
\end{aligned}
\end{equation}

\subsubsection{Optimization of dictionary}

To optimize the dictionary matrix $D$, we remove the terms irrelevant to $D$ from the objective, fix the other variables, and obtain the following optimization problem,

\begin{equation}
\label{equ:obj_d}
\begin{aligned}
\min_{D}
&
\sum_{i=1}^n  \left \| \bfx_i - D \bfs_i \right \|_2^2,\\
s.t.~
&
\|\bfd_j\|_2^2 \leq c, \forall ~j=1, \cdots, m.
\end{aligned}
\end{equation}
The dual optimization problem for this problem is

\begin{equation}
\label{equ:Lag_d}
\begin{aligned}
\max_{\alpha_j|_{j=1}^m}~& \min_{D} \left \{ \mathcal{L}(D,\alpha_j|_{j=1}^m)
=
\sum_{i=1}^n  \left \| \bfx_i - D \bfs_i \right \|_2^2 + \sum_{j=1}^m \alpha_j \left ( \|\bfd_j\|_2^2 - c \right )\right .\\
& \left .=
\sum_{i=1}^n  \left \| \bfx_i - D \bfs_i \right \|_2^2 + \sum_{j=1}^m \alpha_j  \|\bfd_j\|_2^2 - \sum_{j=1}^m \alpha_j  c \right .,\\
& \left .=
\sum_{i=1}^n  \left \| \bfx_i - D \bfs_i \right \|_2^2
+ Tr \left (D diag(\alpha_1, \cdots, \alpha_m) D^\top  \right ) - \sum_{j=1}^m \alpha_j  c \right \},
 \\
s.t.~& \alpha_j \geq 0, j=1, \cdots, m,
\end{aligned}
\end{equation}
where $\alpha_j$ is the Lagrange multiplier for the constrain $\|\bfd_j\|_2^2 \leq c$,  $diag(\alpha_1, \cdots, \alpha_m)$ is a diagonal matrix with ist diagonal elements as $\alpha_1, \cdots, \alpha_m$, and  $\mathcal{L}(D,\alpha_j|_{j=1}^m)$ is the Lagrange function. To minimize the Lagrange function with regard to $D$, we set its gradient with regard to $D$ to zero, and we have

\begin{equation}
\label{equ:Lag_Dder}
\begin{aligned}
&\nabla \mathcal{L}_D=
-2 \sum_{i=1}^n  \left ( \bfx_i - D \bfs_i\right ) \bfs_i^\top  + 2 D diag(\alpha_1, \cdots, \alpha_m) = 0,\\
&\Rightarrow D = \left ( \sum_{i=1}^n \bfx_i \bfs_i^\top \right ) \left ( \sum_{i=1}^n \bfs_i\bfs_i^\top + diag(\alpha_1,\cdots, \alpha_m) \right )^{-1}
\end{aligned}
\end{equation}
To solve the Lagrange multiplier variables, we use the gradient ascent algorithm to obtain $\alpha_1, \cdots, \alpha_m$ in each iteration. After we obtain $\alpha_1, \cdots, \alpha_m$, we can obtain $D$  according to (\ref{equ:Lag_Dder}).

\subsection{Iterative algorithm}

\begin{figure}[!htb]
  \centering
  \includegraphics[width=0.7\textwidth]{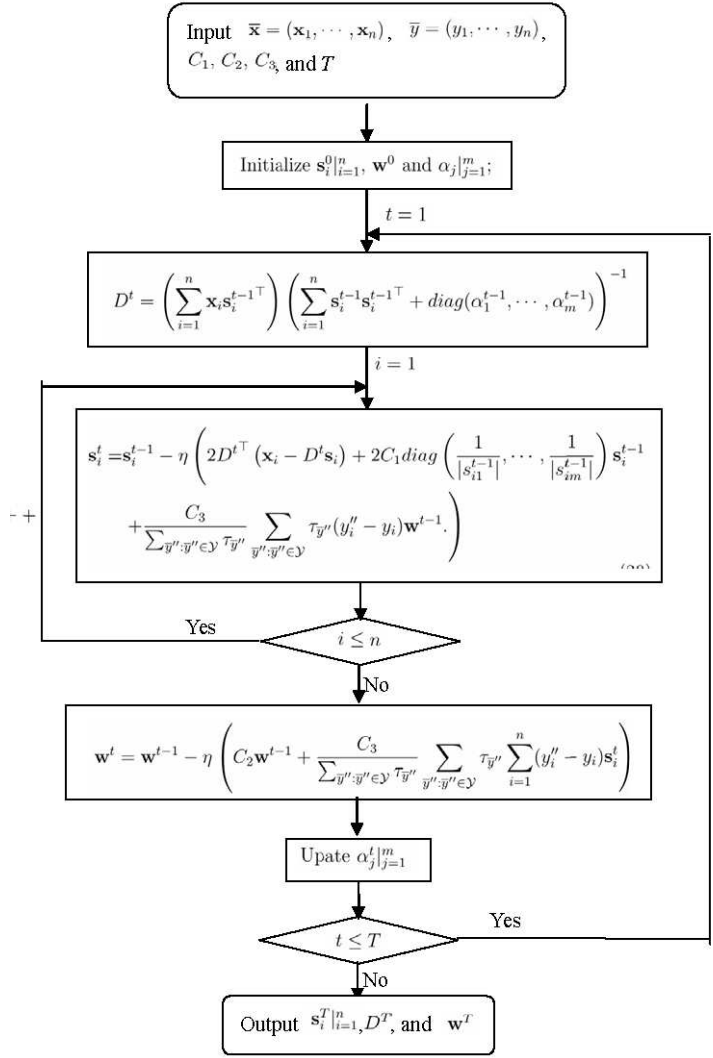}\\
  \caption{The flowchart of the iterative algorithm of JSCHP.}
  \label{fig:FigChartIII150604}
\end{figure}

Based on the optimization results, we develop a novel iterative algorithm, named JSCHP. The algorithm is described in Algorithm \ref{alg:iter}. As we can see from the algorithm, the iterations are repeated $T$ times, and in each iteration, the variables are updated sequentially. The flowchart of the proposed iterative algorithm is given in Fig. \ref{fig:FigChartIII150604}.

\begin{algorithm}[htb!]
\caption{Iterative learning algorithm of joint learning of sparse code and hyper-predictor parameter for multivariate performance measure optimization (JSCHP).}
\label{alg:iter}
\begin{algorithmic}
\STATE \textbf{Input}: A training tuple of $n$ data points $\overline{\bfx} = (\bfx_1, \cdots, \bfx_n)$, and its corresponding class label tuple $\overline{y} = (y_1, \cdots, y_n)$;

\STATE {\textbf{Input}: Tradeoff parameters $C_1$, $C_2$, $C_3$.}

\STATE \textbf{Input}: Maximum iteration number $T$;

\STATE Initialize $\bfs_i^0|_{i=1}^n$, $\bfw^0$ and $\alpha_j|_{j=1}^m$;

\FOR{$t=1,\cdots,T$}

\STATE Update $D^t$ by the updating rule in (\ref{equ:Lag_Dder}) and fixing $\bfs_i^{t-1}|_{i=1}^n$ and $\alpha_j^{t-1}|_{j=1}^m$,

\begin{equation}
\begin{aligned}
D^t = \left ( \sum_{i=1}^n \bfx_i {\bfs_i^{t-1}}^\top \right ) \left ( \sum_{i=1}^n {\bfs_i^{t-1}}{\bfs_i^{t-1}}^\top + diag(\alpha_1^{t-1},\cdots, \alpha_m^{t-1}) \right )^{-1}
\end{aligned}
\end{equation}

\FOR{$i = 1, \cdots, n$}

\STATE Update $\bfs_i^t$ by the updating rule in (\ref{equ:gradient_s}) and fixing $\bfs_i^{t-1}$ and $D^{t}$,

\begin{equation}
\begin{aligned}
\bfs_i^{t}  =
& \bfs_i^{t-1} - \eta
\left (
2 {D^{t}}^\top \left ( \bfx_i - {D^{t}} \bfs_i \right ) + 2 C_1  diag \left (\frac{1}{|s_{i1}^{t-1}|}, \cdots, \frac{1}{|s_{im}^{t-1}|} \right ) \bfs_i^{t-1}
\vphantom{\frac{\sum_{1_2^3}}{1}}
\right .
\\
&
\left .
+
\frac{C_3}{\sum_{\overline{y}'': \overline{y}''\in \mathcal{Y}}
\tau_{\overline{y}''}}
\sum_{\overline{y}'': \overline{y}''\in \mathcal{Y}}
\tau_{\overline{y}''}  (y_i'' - y_i) \bfw^{t-1}.
\right )
\end{aligned}
\end{equation}
\ENDFOR

\STATE Update $\bfw^t$ by the updating rule in (\ref{equ:obj_w2}) and fixing $\bfw^{t-1}$,

\begin{equation}
\begin{aligned}
\bfw^{t} = \bfw^{t-1} - \eta
\left (
C_2\bfw^{t-1}  +
\frac{C_3}{\sum_{\overline{y}'': \overline{y}''\in \mathcal{Y}}
\tau_{\overline{y}''}}
\sum_{\overline{y}'': \overline{y}''\in \mathcal{Y}}
\tau_{\overline{y}''}
\sum_{i=1}^n (y_i'' - y_i)  \bfs_i^{t}
\right )
\end{aligned}
\end{equation}

\STATE Upate $\alpha_j^{t}|_{j=1}^m$ by fixing $\bfs_i^t|_{i=1}^n$ and using gradient ascent;

\ENDFOR

\STATE \textbf{Output}:
The sparse codes $\bfs_i^T|_{i=1}^n$, dictionary matrix $D^T$, and hyper-predictor parameter $\bfw^T$.

\end{algorithmic}
\end{algorithm}

{
The novelty of this algorithm is of three folds:
\begin{enumerate}
\item This algorithm is the first algorithm to learn the sparse codes, dictionary and hyper-predictor jointly.
\item This algorithm is the first algorithm to use gradient descent principle to update the hyper-predictor parameters. Traditional hyper-predictor parameter learning method for multivariate performance optimization is based on solving a quadratic programming problem in each iteration, which is time-consuming. Our algorithm gives up the quadratic programming problem, and instead, we used a simple gradient descent rule to update the parameters efficiently.
\item This algorithm is also the first algorithm to solve the sparse codes using the gradient descent rule. Traditional sparse coding algorithm solve the sparse codes by optimizing a $\ell_2$ norm regularized problem directly, which is not convex and time-consuming. We convert the $\ell_1$ norm regularization to a $\ell_2$ norm regularization, which can be easily solved by gradient descent because it is convex.
\end{enumerate}}

{Please note that the input of the iterative algorithm requires the parameters $C_1$, $C_2$ and $C_3$. $C_1$ is the weight of the sparsity term of the sparse code, $C_2$ is the weight of the model complexity term, and $C_3$ is the weight of the losses over the training set.}

\section{Experiments}
\label{sec:exp}

In this experiment, we evaluate the proposed algorithm and compare it against state-of-the-art multivariate performance optimization methods.

\subsection{Data sets}

In the experiment, we used the following three data sets.

\begin{itemize}
\item \textbf{VANET misbehavior data set}:
The first data set is for the problem of detecting misbehaving network nodes of Vehicular Ad Hoc Networks (VANETs) \cite{Grover2011644}. To construct this data set, we used NCTUns-5.0 simulator to conduct simulations, and collected data of 1395 nodes. These nodes belong to two different classes, which are honest nodes and misbehaving nodes. The number of honest nodes is 837, and the number of the misbehaving nodes is 558. Given a candidate nodes, the problem of misbehavior detection is to predict if is a honest node, or a misbehaving node. Thus this is a binary classification problem. To extract the features from each node, we calculate multifarious features, including speed-deviation of node, received signal strength (RSS), number of packets delivered, dropped packets etc.

\item \textbf{Profile injection attacks data set}: The second data set is for the problem of detecting profile injection attacks in collaborative recommender systems \cite{Zhang201496}.  It is well known that collaborative recommender systems is vulnerable to profile injection attacks. Injection attacks is defined as malicious users inserting fake profiles into the rating database, and biasing the systems' output. To construct the data set, we randomly select 1000 genuine user profiles from Movielens 1M dataset as positive data points, and randomly generate 300 attacking fake user profiles as negative data points. The problem of profile injection attacks detection is to classify a candidate user profile to genuine user or fake user. To extract features from each user profile, we first calculate its rating series  based on the novelty and popularity of items, and then use the empirical mode decomposition (EMD) to decompose its rating series, and finally extract Hilbert spectrum based features.

\item \textbf{UT-kinect 3D action data set}: The third data set if for the problem of recognizing human actions from 3D body data. In this data set, there are 200 3D body data samples, and each 3D body data samples is treated as a data point. These data points belong to 10 different action classes. The number of data points for each class is 20. The 10 classes are listed as follows: walk, sit down, stand up, pickup, carry, throw, push, pull, wave and clap hands \cite{xia2012view}. To extract features from each data point, we calculate the histogram of the 3D joints of each data point.

\end{itemize}

\subsection{Experiment setup}

To perform the experiments, we used the 10-fold cross validation. A data set is split to 10 folds randomly. Each fold was used as a test set in turn. The remaining 9 folds were combined and used as a training set. Given a desired multivariate performance measure, we performed the proposed algorithm on the training set to learn the dictionary and the classifier parameter. Then we used the learned dictionary and the classifier to classify the test data points. Finally, we compared the classification results of the test data points against the true class labels using the given multivariate performance measure.

The following multivariate performance measures were used.

\begin{itemize}

\item \textbf{F1 score}: The first multivariate performance measure is the F1 score, and it is defined as

\begin{equation}
\label{equ:f}
\begin{aligned}
&F1 score\\
&=  \frac{2\times Number~of~correctly~classified~positive~data~points}{
\begin{pmatrix}
2\times Number~of~correctly~classified~positive~data~points\\
+ Number~of~wrongly ~classified~data~points
\end{pmatrix}
}.
\end{aligned}
\end{equation}

\item \textbf{PRBEP}: The third multivariate performance measure is PRBEP,  precision-recall
curve eleven point. It is defined as a point where precision and recall values are equal to each other. The precision-recall curve is obtained by plotting precisions against recalls. Precision and recall are defined as,

\begin{equation}
\label{equ:PRBEP}
\begin{aligned}
&precision =
\frac{Number~of~correctly~classified~positive~data~points}{Nnumber~of~data~points~classified~as~positive},\\
&recall = \frac{Number~of~correctly~classified~positive~data~points}{Total~number~of~positive~test~data~points}.
\end{aligned}
\end{equation}
We can generate different groups of precisions and recalls, and plot precisions against corresponding recalls to obtain the precision-recall curve. The point in the curve where precision is equal to the recall is defined as PRBEP.

\item \textbf{AUC}: The second multivariate performance measure is the AUC, area under operating characteristic curve. Operating characteristic curve is defined as a curve obtained by plotting true positive rate against false positive rate.  True positive rate and false positive rate are defined as follows,

\begin{equation}
\label{equ:auc}
\begin{aligned}
&true~ positive~ rate =  \\ &\frac{Number~of~correctly~classified~positive~data~points}{Total~number~of~positive~test~data~points},\\
&false~ positive~ rate =  \\ &\frac{Number~of~wrongly~classified~negative~data~points}{Total~number~of~negative~test~data~points}.
\end{aligned}
\end{equation}
By changing a threshold parameter of the classifier, we can have different groups of true positive rates and false positive rates. Plotting true positive rates against its corresponding false positive rates, the operating characteristic curve can be obtained.

\end{itemize}

\subsection{Experiment results}

\subsubsection{Comparison to state-of-the-arts}

\begin{figure}
  \centering
\subfigure[$F1 score$]{
  \includegraphics[width=0.3\textwidth]{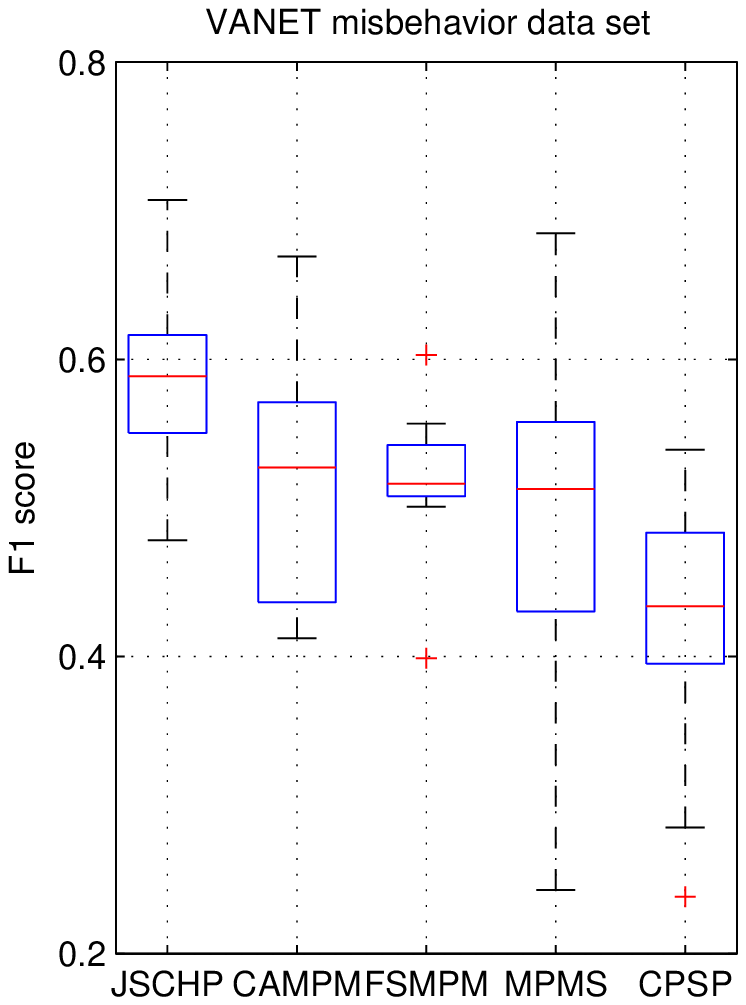}}
\subfigure[PRBEP]{
  \includegraphics[width=0.3\textwidth]{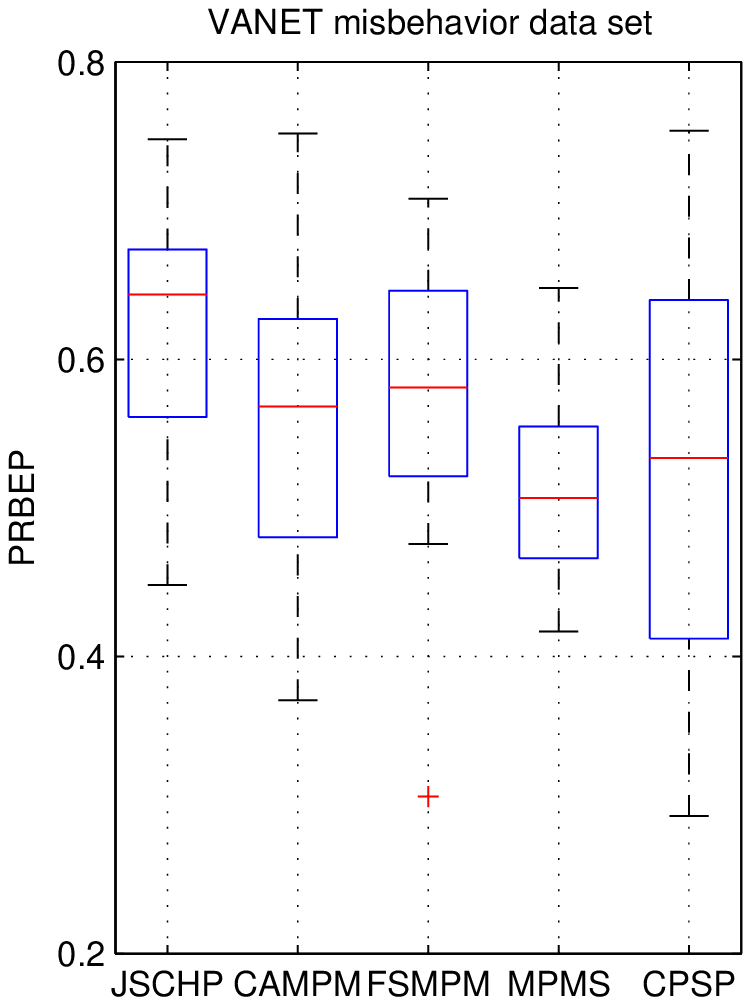}}
\subfigure[AUC]{
  \includegraphics[width=0.3\textwidth]{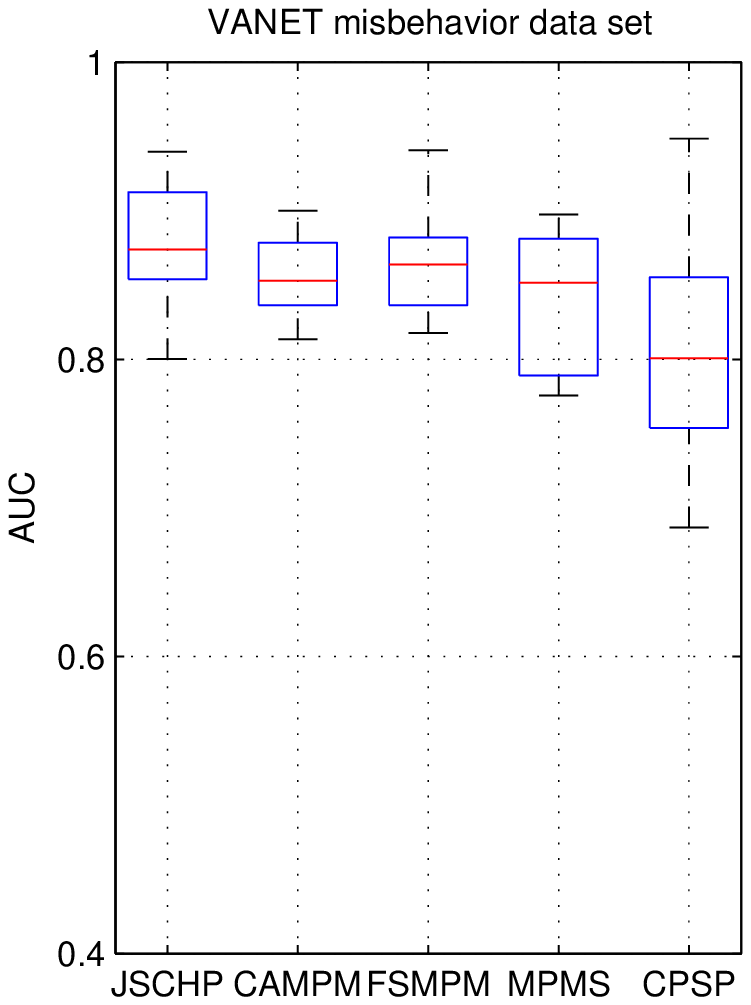}}
  \\
  \caption{Results of comparison to state-of-the-arts on VANET misbehavior data set.}
  \label{fig:VANET}
\end{figure}

\begin{figure}
  \centering
\subfigure[$F1 score$]{
  \includegraphics[width=0.30\textwidth]{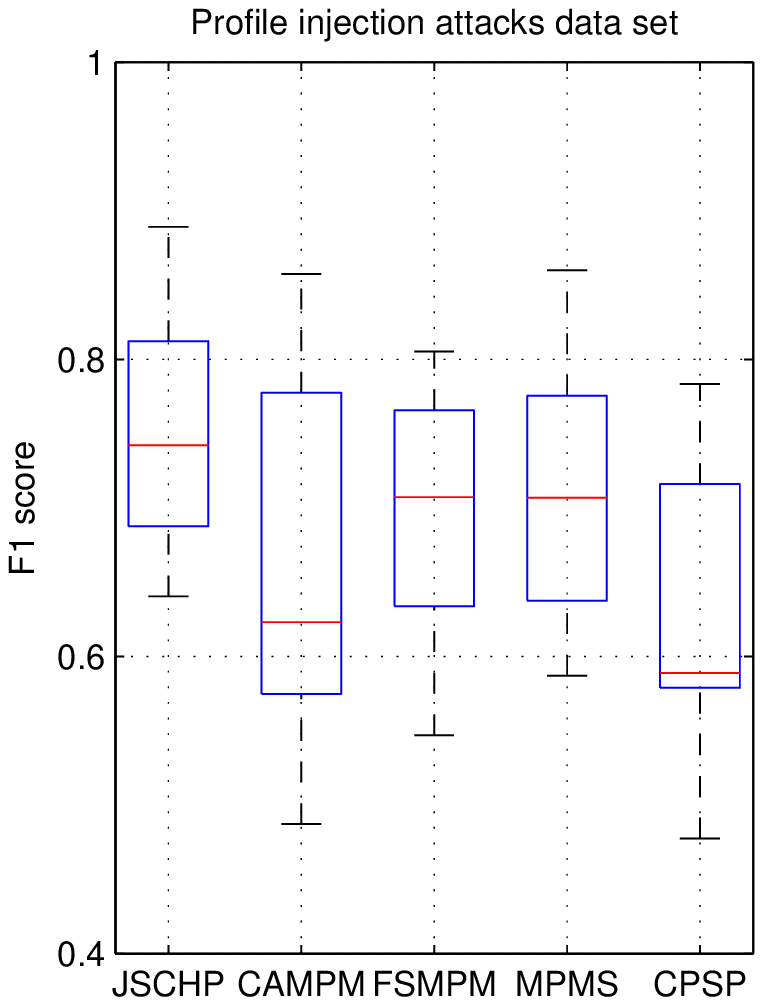}}
\subfigure[PRBEP]{
  \includegraphics[width=0.30\textwidth]{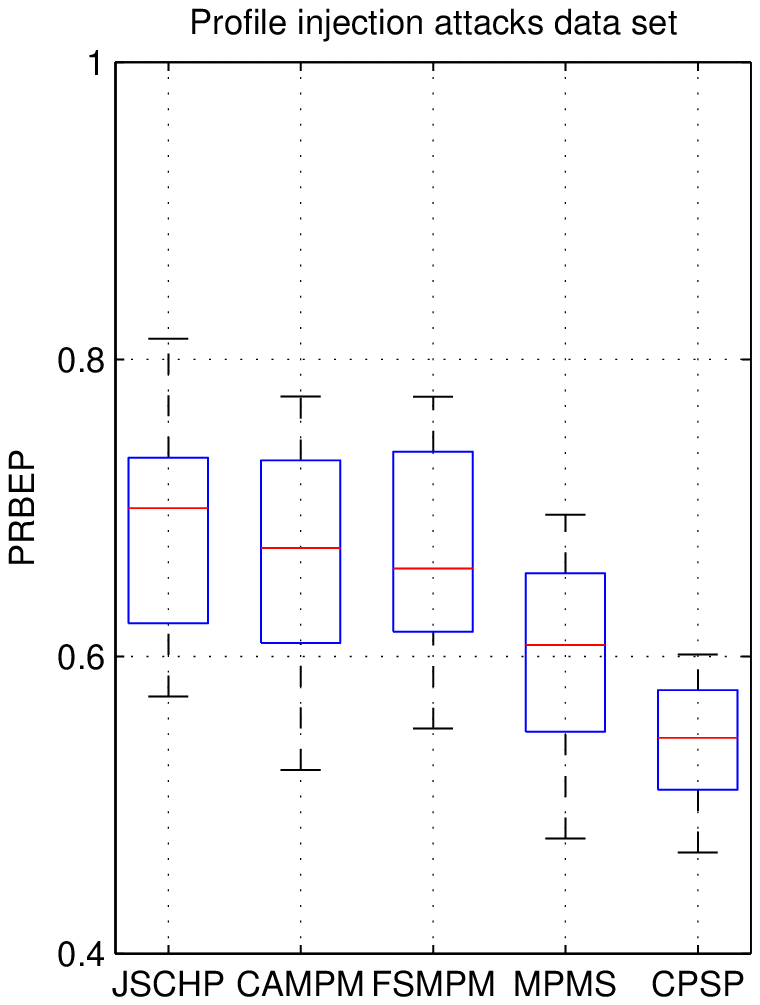}}
\subfigure[AUC]{
  \includegraphics[width=0.30\textwidth]{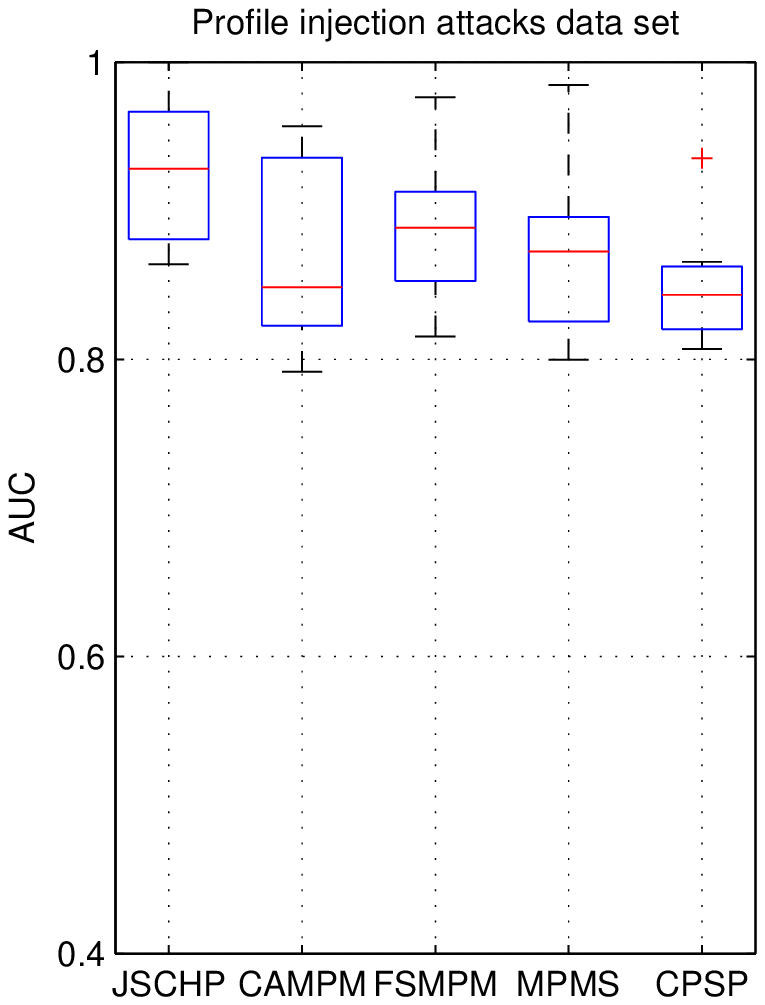}}
  \\
  \caption{Results of comparison to state-of-the-arts on profile injection attacks data set.}
  \label{fig:Attack}
\end{figure}

\begin{figure}
  \centering
\subfigure[$F1 score$]{
  \includegraphics[width=0.30\textwidth]{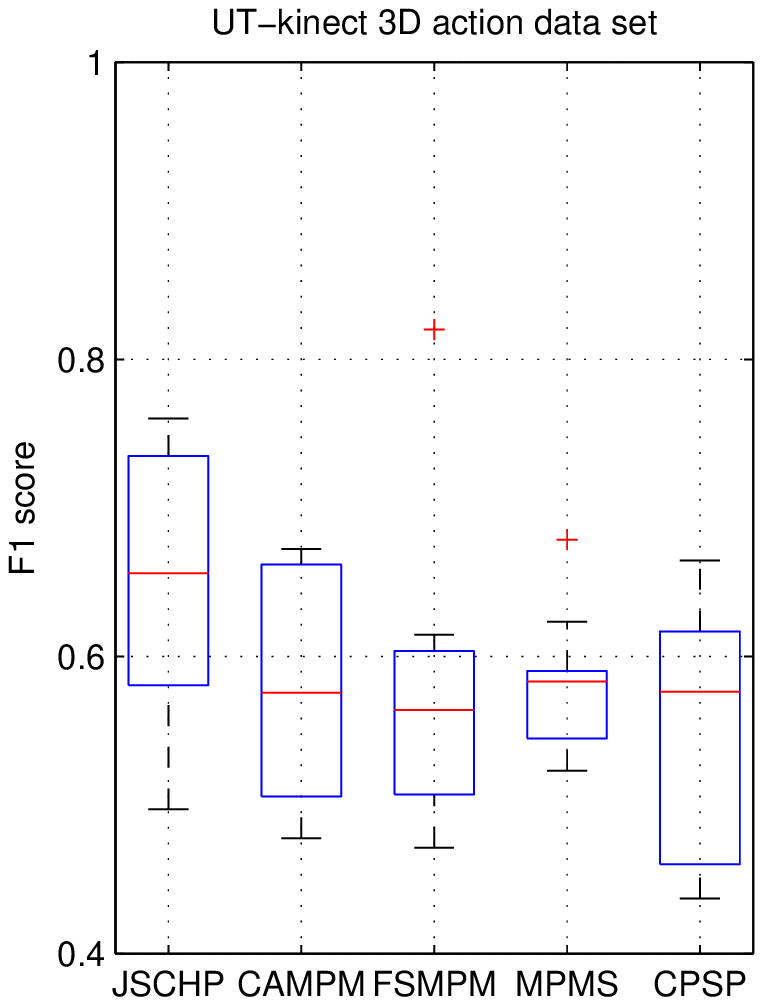}}
\subfigure[PRBEP]{
  \includegraphics[width=0.30\textwidth]{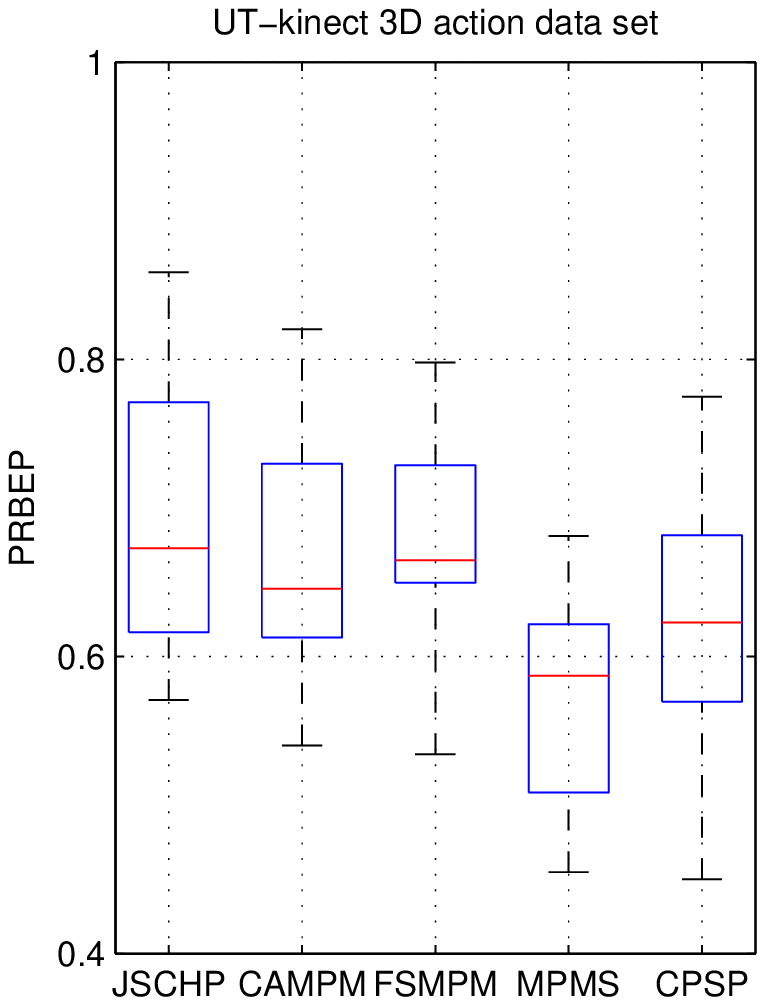}}
\subfigure[AUC]{
  \includegraphics[width=0.30\textwidth]{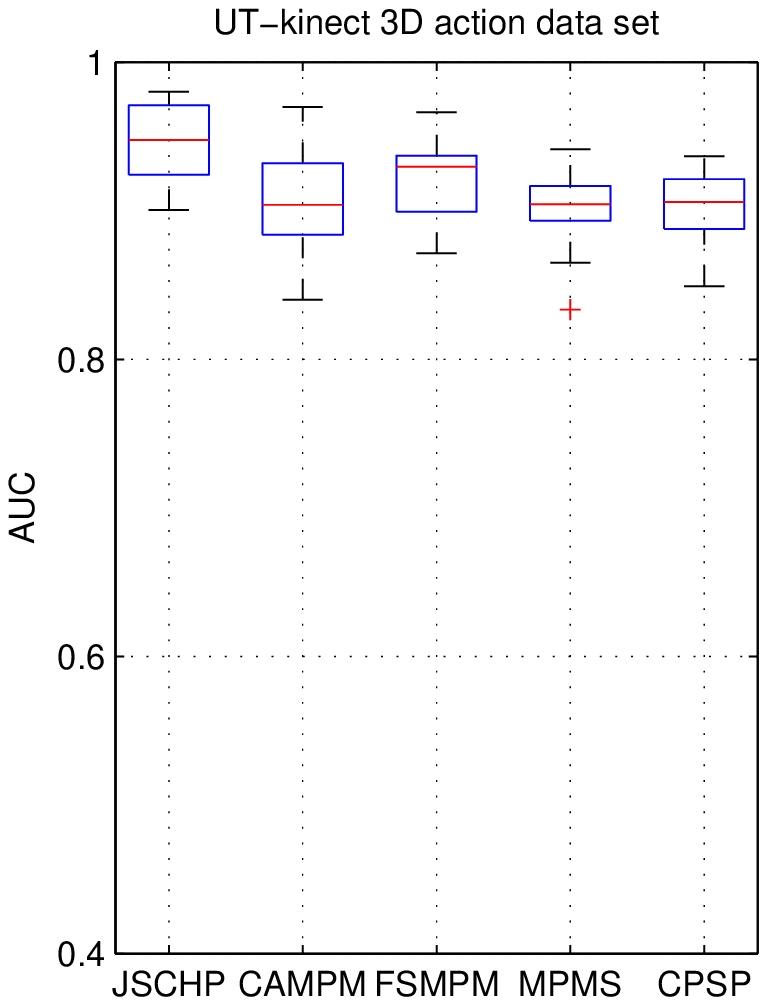}}
  \\
  \caption{Results of comparison to state-of-the-arts on UT-kinect 3D action data set.}
  \label{fig:UT}
\end{figure}

In this experiment, we first compared the proposed algorithm JSCHP to some state-of-the-art machine learning algorithms for multivariate performance optimization, including the cutting-plane subspace pursuit (CPSP) \cite{Joachims2005377},  multivariate performance measure smoothing (MPMS) \cite{Zhang20123623}, feature selection based multivariate performance measure optimization (FSMPM) \cite{Mao20132051}, and classifier adaptation based multivariate performance measure optimization (CAMPM) \cite{Li20131370}. The boxplots of different performances measures of the 10-fold cross validation over different data sets are given in Fig. \ref{fig:VANET}, \ref{fig:Attack} and \ref{fig:UT}. From these figures, we can see that the proposed algorithm JSCHP outperforms the compared algorithms in most cases. For example, in the experiments over VANET misbehavior data set, when  PRBEP performance is considered, only JSCHP algorithm achieves a median value higher than 0.6, while  the media values of all other algorithms are lower than 0.6. Moreover, in the experiments over UT-kinect 3D action data set, we can see that the median value of the $F1$ scores of JSCHP is even higher than the 75-th percentile values of other algorithms. These are strong evidences that the proposed algorithm is more effective than the compared algorithms for the problem of optimizing multivariate performance measures. It is also interesting to see that AUC seems a easier multivariate performance measure to optimized than $F1$ score and PRBEP. In all the experiments over three data sets, the observed AUC values are higher then corresponding $F1$ scores and PRBEP values. The results of CAMPM, FSMPM and MPMS are comparable to each other, and better than CPSP.

\subsubsection{Parameter sensitivity}

\begin{figure}
\centering
\subfigure[$C_1$ and $C_2$]{
\includegraphics[width=0.47\textwidth]{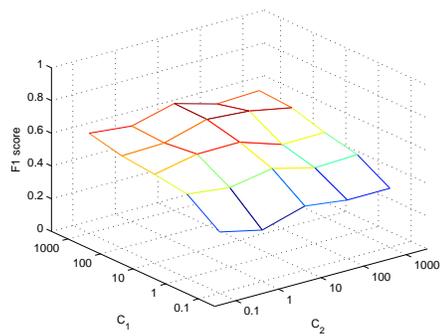}
\label{fig:c1}}
\subfigure[$C_1$ and $C_3$]{
\includegraphics[width=0.47\textwidth]{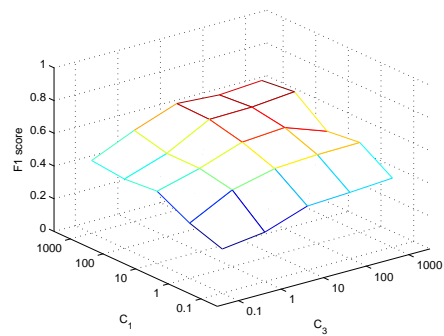}
\label{fig:c2}}
\subfigure[$C_2$ and $C_3$]{
\includegraphics[width=0.47\textwidth]{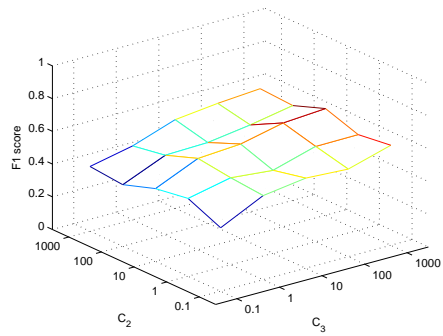}
\label{fig:c3}}
\\
\caption{Parameter sensitivity curves of $F_1$ scores over UT-kinect 3D action data set.}
\label{fig:Param}
\end{figure}

We are also interested in the sensitivity of the proposed algorithm against three tradeoff parameters $C_1$, $C_2$ and $C_3$. Thus we varied the tradeoff parameters $C_1$, $C_2$ and $C_3$ contemporaneously to compute the sensitivity of the algorithm to the parameters. The average $F_1$ score of the proposed algorithm of combinations of different values of these parameters are given in Fig. \ref{fig:Param}. From Fig. \ref{fig:c1} and Fig. \ref{fig:c2}, we can see that when $C_1$ is increasing, the performances are also being improved. $C_1$ is the weight of the sparsity term of the sparse code, and from the experiment results, we can conclude that when we have a larger sparsity penalty, the performance can be better. This means that a sparse representation is important for learning hyper-predictor to optimize multivariate performance measures. It is well known that sparse representation can benefit the learning of a good classifier using common and simple performance measures. However, it is still unknown if such sparse representation can also benefit the learning of hyper-predictor for complex multivariate performance measure optimization. Our experiments answer this question, and we find that the sparsity of the presentation is also important for the optimization of complex multivariate performance measures, just like it works for the simple performance measure optimization. From Fig. \ref{fig:c1} and Fig. \ref{fig:c3}, we can see that the improvement of the performances against the $C_2$ parameter is not clear. However, the performance is stable for different parameters. This parameter is the weight for the complexity of the hyper-predictor parameter. From the results, we cannot conclude that a simpler predictor can optimize the multivariate performance measure better than a complex predictor.
From Fig. \ref{fig:c2} and Fig. \ref{fig:c3}, we can see that a larger $C_3$ can also improve the performance. This is because $C_3$ is the weight of the upper bound of the corresponding loss function. A larger $C_3$ can lead to a better solution for the minimization of the loss function, and thus leads to a better performance measure.

\subsubsection{Running time}

\begin{figure}
  \centering
  \includegraphics[width=0.6\textwidth]{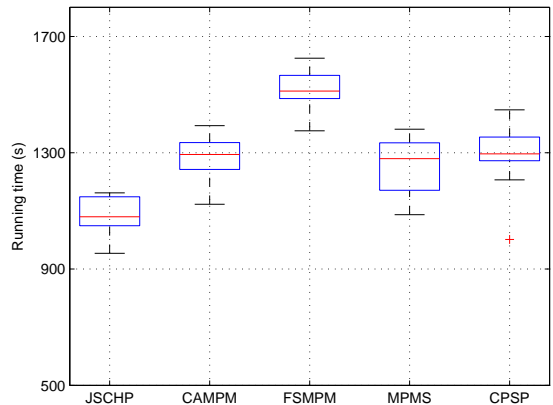}\\
  \caption{Boxplots of running time of 10-fold cross validation over UT-kinect 3D action data set.}
  \label{fig:time}
\end{figure}

We are also interested in the running time of the proposed algorithm and the compared algorithms. The boxplots of running time of different algorithms of the 10-fold cross validation over UT-kinect 3D action data set is given in Fig. \ref{fig:time}. It is obvious that the proposed algorithm has shorter running time than the other algorithms. A possible reason is that the other algorithms are based on cutting-plane algorithm. In this algorithm, in each iteration, a active set is maintained, and a quadratic programming algorithm is solved over this active set. The solving of the quadratic algorithm is time consuming. Moreover, to update the active set, we need to seek a maximization over all possible class label tuples. However, in our algorithm, we only seek a maximization in the class label tuple space to approximate the upper bound, and no quadratic programming problem is considered, while only a gradient descent updating procedure is conducted.

\section{Conclusion and future works}
\label{sec:conclud}

In this paper, we proposed a novel method for the problem of multivariate performance measure optimization. This method is based on joint learning of sparse codes of data point tuple and a hyper-predictor to predict the class label tuple. In this way, the sparse code learning is guided by the minimization of the multivariate loss function corresponding to the desired multivariate performance measure. Moreover, we also proposed a novel upper bound approximation of the multivariate loss function. We model the learning problem as an minimization problem and solve it by developing a iterative algorithm based on gradient descent method. The proposed algorithm is compared to state-of-the-art multivariate performance measure optimization algorithms, and the results show its advantage. In the future, we will consider extend the proposed framework to structured label prediction problem, since it is similar to multivariate performance measure optimization. In the future, we will also use the proposed algorithm for the application of computer vision \cite{wang2015supervised,wang2015image}.


\end{document}